%% file: acl_latex.tex
\pdfoutput=1

\documentclass[11pt]{article}

\usepackage[preprint]{acl}

\usepackage{times}
\usepackage{latexsym}

\usepackage[T1]{fontenc}

\usepackage[utf8]{inputenc}

\usepackage{microtype}

\usepackage{inconsolata}

\usepackage{graphicx, multirow}
\usepackage{todonotes}
\usepackage{enumitem}
\usepackage{subcaption}

\usepackage{booktabs}

\usepackage{multirow}
\usepackage{tabularx}
\usepackage{array}
\usepackage{makecell}
\usepackage{float}
\usepackage{algorithm}
\usepackage{algorithmic}
\usepackage{amsmath}

\newcommand{\App}{EventFull}

\title{\textit{\App{}}: Complete and Consistent Event Relation Annotation}

\author{Alon Eirew \quad
        Eviatar Nachshoni \quad 
        Aviv Slobodkin \quad 
        Ido Dagan \\
        Bar-Ilan University \\ {\tt \{alon.eirew, eviatarn, lovodkin93\}@gmail.com} \\ {\tt dagan@cs.biu.ac.il}}

\begin{document}
\maketitle

\input{sections/abstract}

\input{sections/intro}
\input{sections/background}

\input{sections/tool}

\input{sections/pilot}

\input{tables/annot_time}
\input{tables/annot_steps}

\input{sections/conclusion}
\bibliography{acl_latex}

\appendix

\input{sections/appendix}

\end{document}

%% file: sections/abstract.tex
\begin{abstract}
Event relation detection is a fundamental NLP task, leveraged in many downstream applications, whose modeling requires datasets annotated with event relations of various types. However, systematic and complete annotation of these relations is costly and challenging, due to the quadratic number of event pairs that need to be considered. Consequently, many current event relation datasets lack systematicity and completeness.
In response, we introduce \textit{\App{}}, the first tool that supports consistent, complete and efficient annotation of temporal, causal and coreference relations via a unified and synergetic process.
A pilot study demonstrates that \App{} accelerates and simplifies the annotation process while yielding high inter-annotator agreement.\footnote{The tool and code are publicly available at \url{https://github.com/AlonEirew/EventGraphAnnot}.}





\end{abstract}

%% file: sections/intro.tex
\section{Introduction}
\label{sec:intro}
Identifying the semantic relations between events mentioned in a text, notably temporal, causal and coreference relations, has been a fundamental goal in NLP.
Substantial efforts have been devoted to developing various datasets that capture some or all of these relations \cite{ogorman-etal-2016-richer, hong-etal-2016-building, wang-etal-2022-maven}. These datasets were then leveraged to develop and to evaluate corresponding models for detecting event-event relations \cite{Hu2023ProtoEMAP, Guan2024TacoERECC}. 
The output of such models has been utilized in a range of downstream applications, with recent examples including event forecasting \cite{Ma2023ContextawareEF}, misinformation detection \cite{lei-huang-2023-identifying}, and treatment timeline extraction \cite{yao-etal-2024-overview}, among others.

\begin{figure*}[t!]
\centering
\includegraphics[width=\textwidth]{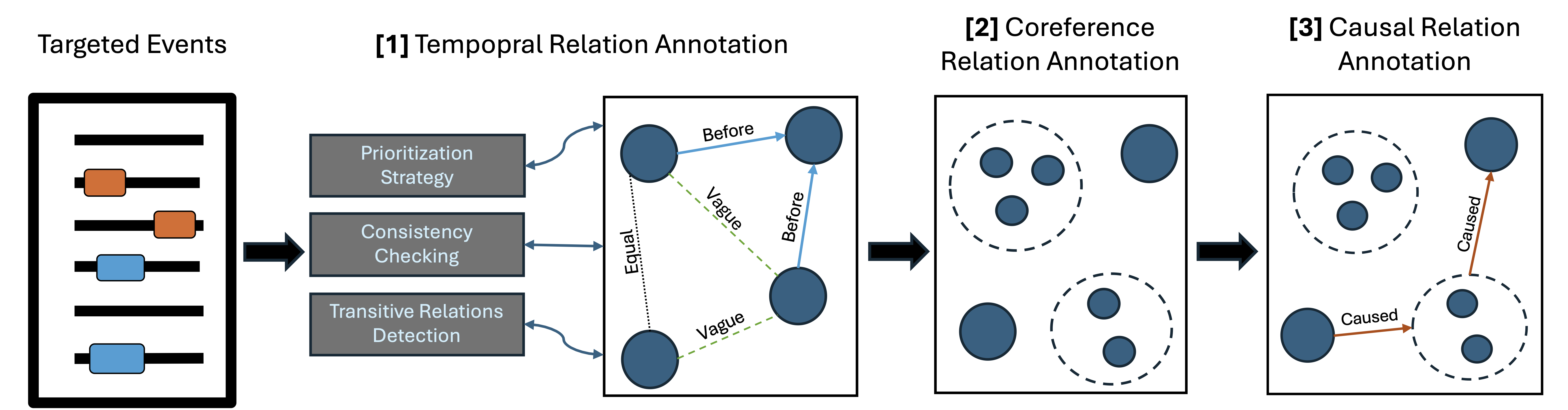}
\caption{The \App{} annotation pipeline begins with a document containing marked targeted events and proceeds through three stages: [1] \textbf{Temporal Relations:} annotators establish temporal relations between event pairs, supported by three processes, including pair prioritization strategy, consistency checking, and transitive relation detection (§\ref{sec:tool-flow}); [2] \textbf{Coreference:} annotators identify coreferring event mentions; [3] \textbf{Causal Relations:} annotators determine causal relations for pairs of events.}
\label{fig:figure1}
\end{figure*}



Models for detecting event relations are expected to produce \textit{complete} output, that is identifying \textit{all} event relations that can be inferred from the given text. 
Accordingly, such models should ideally be evaluated, and trained, on datasets in which event relation annotation is in itself complete, in the sense that each pair of targeted events has been classified for its potential relationships. 
Unfortunately, though, exhaustive manual annotation of all event-event relations is typically considered extremely challenging or impractical, since the number of event pairs to be considered is quadratic in the number of targeted event mentions in the text \cite{naik-etal-2019-tddiscourse, rogers2022narrativetime}. As the number of event mentions grows, this task quickly becomes both too time consuming and cognitively unmanageable.


Faced with this inherent annotation complexity, many datasets adopted an annotation protocol that restricts the number of events or event pairs considered for annotation through various restrictions and heuristics. Notably, TB-Dense \cite{chambers-etal-2014-dense}, ECB+ \cite{cybulska-vossen-2014-using}, MEANTIME \cite{minard-etal-2016-meantime}, and EventStoryLine \cite{caselli-vossen-2017-event} restrict event pairs to a span of two consecutive sentences. This limitation inherently prevents testing and training models on longer-range relations.
Other datasets, such as TimeBank \cite{pustejovsky-etal-2003} and MAVEN-ERE \cite{wang-etal-2022-maven}, did not publish a systematic annotation execution protocol that guarantees actual complete annotation, and were subsequently criticized for being incomplete in their relation annotation \cite{pustejovsky-stubbs-2011-increasing, rogers2022narrativetime}.
Further, some researchers aimed to avoid the cost of manual annotation altogether and employed fully- or partly-automatic dataset creation methods \cite{mirza-etal-2014-annotating, madaan-yang-2021-neural, tan-etal-2024-set}. These approaches inherently incorporate biases introduced by the employed automated method, making them less reliable for testing purposes while being prone to yield biased models when used for training. 
Finally, motivated by similar observations to ours, the recent NarrativeTime project \cite{rogers2022narrativetime} does emphasize relation annotation completeness, supported by a corresponding annotation tool. However, their work addresses only temporal relations, while employing a complex annotation scheme that requires expert annotators. Indeed, their actual annotation was performed by two of the authors, overall limiting the replicability of this approach.



In this paper, we aim to close major gaps in prior annotation protocols. To that end, we introduce a simple-to-use annotation tool that facilitates future creation of event relation datasets (for covering multiple relation types, new genres and languages, etc.), which \textit{guarantees} complete relation annotation . 
Our web-based tool, \textit{\App{}} (Figure \ref{fig:figure1}), fulfills four critical goals, which constitute our primary contributions: (1) supporting joint synergetic annotation of the three prominent event-event relation types: temporal, causal, and coreference; (2) for any given set of targeted events within the text, the gold output relations are guaranteed to be \textit{complete}, classifying all event pairs for the three relation types; (3) supporting efficient annotation while guaranteeing annotation consistency, via automated transitive completion and consistency checks, while leveraging constraints imposed across the three relation types (see \S\ref{sec:tool-flow}); (4) ease of use by a non-expert annotator, lowering the bar for future dataset creation. 
To the best of our knowledge, \App{} is the first available tool that effectively supports and integrates all these goals.

In the remainder of the paper, we first survey relevant background and related work (\S\ref{sec:background}), followed by the presentation of \App{}'s input and output structure and its workflow (\S\ref{sec:tool}). Finally, we present a pilot study in which we demonstrate the effectiveness of \App{} (\S\ref{sec:pilot}).

%% file: sections/background.tex
\section{Background and Related Work}
\label{sec:background}
This section first provides relevant background regarding event relations, followed by a short survey of prior annotation tools.

\subsection{Event Relations}
\label{sec:background-relations}
We focus on the three event-event relations which have been addressed most broadly in the literature: \textbf{temporal} \cite{do-etal-2012-joint} --- identifying the temporal order between events; \textbf{coreference} \cite{raghunathan-etal-2010-multi} --- indicating whether two event mentions in the text refer to the same real-world event; and \textbf{causal} \cite{mirza-etal-2014-annotating} --- detecting whether one event caused another event to happen.\footnote{Another relation type often considered is \textit{sub-event} \cite{glavas-etal-2014-hieve}, which falls beyond the scope of this work.}

Event relations satisfy various constraints. Temporal relations, which connect \textit{any} two events, must maintain transitive consistency \cite{marc-et-al-2005-temporal}: if event A precedes B and B precedes C, then A necessarily precedes C \cite{allen-et-al-1983}. Temporal order induces further constraints on the other two relations. In a causal relation, the causing event must \textit{precede} the other \cite{caselli-vossen-2017-event}, while coreferring event mentions must temporally \textit{co-occur} \cite{cybulska-vossen-2014-using}. As described in \S\ref{sec:tool-flow}, we leverage these constraints in our tool to increase annotation efficiency while guaranteeing its consistency.

Temporal and causal relations have been modeled at varying levels of granularity. The number of temporal relation types ranged from 13-14 in Allen's interval algebra \cite{ALLEN1984123} and TimeML \cite{pustejovsky2003timeml} to 4-6 in recent approaches \cite{chambers-etal-2014-dense, ning-etal-2018-multi, wang-etal-2022-maven}, aiming to increase annotation consistency and scalability. Fine grained causal relations distinguish three sub-types --- \textit{cause}, \textit{precondition}, and \textit{prevention} \cite{ogorman-etal-2016-richer, caselli-vossen-2017-event}, while other approaches model only a single \textit{cause} relation \cite{do-etal-2011-minimally, ning-etal-2018-joint}, which is easier to annotate and model. Adopting fine-grained relations has proven challenging even for expert annotators, as evidenced by low inter-annotator agreement \citep{hong-etal-2016-building, ogorman-etal-2016-richer}. To support annotation by non-experts, we focus on the more coarse-grained relation types (\S\ref{sec:tool-in-out}). 

Finally, we note that various methods have been proposed for determining the set of event mentions over which event relations will be annotated. These include considering all mentions \cite{ogorman-etal-2016-richer}, only verbal mentions \cite{chambers-etal-2014-dense}, only actually-occurring events \cite{ning-etal-2018-multi, wang-etal-2022-maven}, or salient events in the text \cite{madaan-yang-2021-neural, tan-etal-2024-set}. To allow flexibility in the selection of events, we leave event mention selection to be performed independently by dataset creators as a preprocessing step for \App{} (\S\ref{sec:tool-in-out}).

\subsection{Prior Annotation Tools}
\label{sec:background-tools}

Event relation annotation has been carried out by two types of tools. 
The first involves adapting \textit{general-purpose} annotation tools, such as BRAT \cite{stenetorp-etal-2012-brat} and CAT \cite{bartalesi-lenzi-etal-2012-cat}. However, these general tools do not leverage specific properties of event relations, as mentioned above in \S\ref{sec:background-relations},
to support annotation completeness, consistency and efficiency.
In contrast, targeted annotation tools have been developed to support the annotation of individual relations, including for temporal \cite{derczynski-etal-2016-gate, rogers2022narrativetime} and coreference relations \cite{bornstein-etal-2020-corefi}.
Some of these tools are rather complex, suitable for expert annotators. 
\App{}, on the other hand, supports synergetic annotation of all the primary three event relation types, suitable for non-expert annotators, and provides targeted automated support for completeness, consistency and efficiency.


%% file: sections/tool.tex
\section{The \App{} Annotation Tool}
\label{sec:tool}
As described in \S\ref{sec:intro}, \App{} guarantees \textit{complete} and \textit{consistent} annotation for temporal, causal and coreference relations between events mentioned in an input text,  while minimizing the manual annotation effort. We next describe its input and output structure, followed by a detailed description of its functionality and workflow.\footnote{A live demo of \App{} is available at \url{https://eventgraphannotnew.onrender.com}.}


\subsection{Input and output}
\label{sec:tool-in-out}
\paragraph{Input} 
\App{} receives as input a text, marked with targeted event mentions to be covered by the event relation annotation. As mentioned earlier (\S\ref{sec:background-relations}), there are many different approaches for detecting event mentions in a text and for selecting those for which relations will be annotated. Therefore, we intentionally leave the detection and selection of event mentions orthogonal to \App{}, leaving dataset creators the flexibility to choose their own methods for this preprocessing step. Thus, \App{} focuses on the challenging task of producing a complete event relation annotation all input event mentions. 

Nevertheless, as an optional auxiliary functionality, \App{} supports annotators in reviewing the set of marked event mentions in the text, allowing them to filter out some of them as desired (illustrated in Appendix~\ref{appx:additional_printscreens}, Figure~\ref{fig-appx:select}). 

\paragraph{Output}
As discussed in §\ref{sec:background-relations}, in order to support simplified and consistent annotation, we focus on the most prominent relation classes for each relation type. For temporal relations, \App{} adopts the \textit{before}, \textit{after}, \textit{equal}, and \textit{uncertain} relations, following the MATRES dataset \cite{ning-etal-2018-multi}. For coreference, we annotate the \textit{coreference} relation, as defined in ECB+ \cite{cybulska-vossen-2014-using}. For causal relations, we focus exclusively on the \textit{cause} relation, consistent with \citet{do-etal-2011-minimally} and \citet{ning-etal-2018-joint}.

Accordingly, the annotation output first specifies a set of event \textit{coreference-clusters}, where each cluster includes a set of event mentions (possibly a singleton) that refer to the same real-world event, hence providing a representation for that event. Then, each ordered pair of coreference clusters is associated with a \textit{temporal} relation, and, if applicable, with a \textit{cause} relation. Importantly, the annotation process guarantees that each pair of event mentions has been classified by the coreference relation, while each pair of events (coreference clusters) has been classified for their temporal and causal relation.

\subsection{Relation Annotation Workflow}
\label{sec:tool-flow}

\App{} employs a layered annotation approach, dividing the task into three subsequent sub-tasks (screenshots shown in Appendix~\ref{appx:additional_printscreens}), namely temporal, coreference and causal relation annotation, in this order. Each step builds upon the previous one, while leveraging the temporal constraints imposed across the relations (discussed in \S\ref{sec:background-relations}). Each task is supported by guidelines and instructions provided within the tool to assist annotators in performing the tasks (see examples in Appendix~\ref{appx:guidelines}). 
At any stage, annotators can save their progress or revise prior annotations. Once all tasks are completed the system checks for overall annotation completeness and annotations are exported. Next, we provide a detailed description of these steps.

\subsubsection{Temporal Relation Annotation}
The first step aims to establish a temporal relation for \emph{all} pairs of input event mentions (see Figure~\ref{fig-appx:temporal} in Appendix~\ref{appx:additional_printscreens} for illustration).\footnote{Notice that while temporal relations would eventually be induced at the level of coreference clusters, it is overall more efficient to conduct the temporal annotation in the first stage, at the event mention level (see \S\ref{sec:tool-coref} and Table \ref{tab:annot_step}).} Annotators are presented with each pair of event mentions at a time, with the pair under scrutiny highlighted in its text context.
Alongside the text, a graph visualization is provided, with the scrutinized pair connected by an emphasized red edge. Annotators must choose one of four label options for the edge connecting the two events, corresponding to the set of \App{}'s temporal relations (specified in \S\ref{sec:tool-in-out}): \textit{before}, \textit{after}, \textit{equal}, or \textit{uncertain} --- when the temporal relation cannot be inferred from the text. 
Notably, the graph visualization updates with each selection, aiding in tracking progress and providing flexibility by allowing annotators to directly select and annotate (or revise) the relation for event mention pairs by clicking on the corresponding pairs of graph nodes.

\begin{figure}[t!]
\centering
\includegraphics[width=0.48\textwidth]{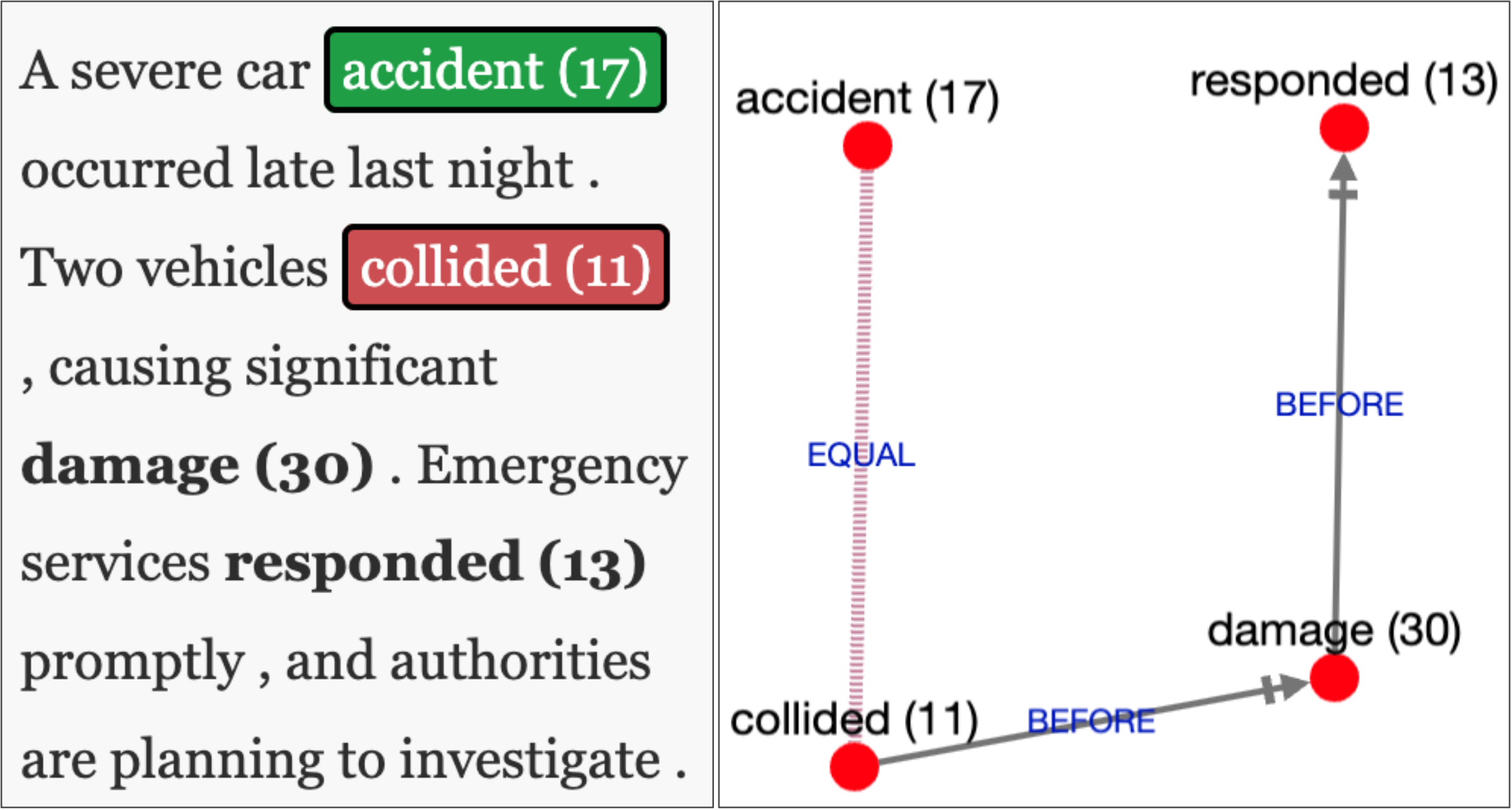}
\caption{A simple example illustrating the prioritization strategy and automatic annotation of transitive relations. The prioritization strategy incrementally presents the pairs \textit{`accident-collided'}, \textit{`collided-damage'}, and \textit{`damage-responded'} based on the relations that the annotator selected in each turn. The transitive relations \textit{`accident-damage'}, \textit{`accident-responded'}, and \textit{`collided-responded'} are detected automatically.}
\label{fig:example}
\end{figure}

To ensure complete and consistent annotation while reducing manual annotation complexity, \App{} incorporates three independent background processes to monitor and adjust annotations (Figure~\ref{fig:figure1}):
(1) a \textit{transitive closure algorithm} \cite{ALLEN1984123}, inspired by the temporal constraints of \citet{ning-etal-2018-joint} (see Appendix~\ref{appx:temporal-constraints}, Table~\ref{tab:disc_temp}), is applied after each annotation. This algorithm identifies and \textit{auto}-annotates yet not-annotated pairs whose links can be transitively deduced from already annotated pairs (see example in Figure~\ref{fig:example}), thereby reducing the number of pairs that require manual annotation;(see Appendix~\ref{appx:transitive-algo} for details.)
(2) To enhance the effectiveness of the above transitive closure algorithm, we leverage the observation that the \textit{before} relation is the most frequent in prior temporal datasets \cite{kishore-he-2024-unveiling}. To that end, we developed a \textit{prioritization strategy} that simulates a Depth-first style search to determine the next pair to present to the annotators, starting with the first event in the text (as illustrated in Figure~\ref{fig:example}). This process is triggered after each relation annotation, streamlining the annotation process while enhancing the utility of the auto-discovery of transitive relations (see Appendix~\ref{appx:prior-strategy} for details).
(3) To ensure annotations are free of conflicts (consistent), a \textit{consistency checking} algorithm is applied right after the transitive closure algorithm, detecting any conflicts that arise, for example --- annotating the relation for \textit{`accident-damage'} as anything other than \textit{before} (Figure~\ref{fig:example}). In such cases, the tool highlights the detected conflicts and notifies the annotator, pointing at the path that led to the conflict for review (see Appendix~\ref{appx:discrepancy} for details).

Once all event mention pairs are annotated without conflicts, a notification confirms that no additional pairs require annotation.

\subsubsection{Coreference Annotation}
\label{sec:tool-coref}
In this step annotators are asked to identify coreferring pairs of event mentions (which refer to the same real-world event; See Figure~\ref{fig-appx:coref} in Appendix~\ref{appx:additional_printscreens} for illustration). 
Inherently, this step requires considering only pairs of event mentions whose temporal relation is \textit{equal} (co-occurring mentions), which immensely reduces the number of pairs that need to be considered (see Table~\ref{tab:annot_step}), while leveraging the preceding temporal annotation step.

This step is guided by the tool, which presents to the annotator a targeted event at a time, proceeding by the text order. For each targeted event, the tool highlights all events that temporally co-occur with it, in both the text as well as in a graphical visualization of all event mentions. The annotator then selects, out of these co-occurring mentions, all those that are judged to corefer with the targeted mention. This selection defines a coreference cluster, and the process iterates for the next available mention in the text. Upon completion, event mentions not included in any cluster are marked as singleton clusters. Additionally, a discrepancy detection algorithm ensures consistency by notifying annotators of any event mention that was linked to distinct clusters, asking them to resolve the conflict.

It should be noted at this point that the consistency checks in the temporal annotation phase (Table \ref{tab:disc_temp}) guarantee that the temporal relations for all co-occurring mentions would be identical, which in turn applies to all mentions in a corefernce cluster.
This induces a consistent temporal relation annotation  at the level of coreference clusters.


\subsubsection{Causal Relation Annotation}
Given the coreference clusters from the previous step, each representing a single real-world event, causal relation annotation now considers pairs of events rather than pairs of event mentions. 
For this step we assessed two prior annotation flows in the literature. The first is the RED protocol \cite{ogorman-etal-2016-richer}, which requires considering independently a causal relations for each pair of events at a time. 
The second is the EventStoryLine methodology \cite{caselli-vossen-2017-event}, where an event is presented alongside \emph{all} its temporally preceding events, out of which the annotator is asked to identify all causing events. As we found in our preliminary experiments, this latter approach is faster, reduces cognitive load and supports higher-quality annotation (details provided in Appendix~\ref{app:cause-scheme-expr}), and hence we adopted it for our annotation flow.


A screenshot of causal relation annotation is illustrated in  Figure~\ref{fig-appx:causal} in Appendix~\ref{appx:additional_printscreens}. 
For each targeted event at a time, annotators review a representative mention of it alongside representative mentions of all its preceding events, which are highlighted in the text and in a visualized graph representation, similar to the previous steps. The annotator is then asked to select which of the preceding events caused the current targeted one, and proceeds to the next targeted event. Following the EventStoryLine methodology, causal relations are considered as independently localized for each pair of events, without induced transitivity, hence not requiring transitive consistency checks.




%% file: sections/pilot.tex
\section{Pilot Study}
\label{sec:pilot}

To evaluate \App{}'s effectiveness, we conducted a small-scale study to assess the time, effort and quality of the resulting annotations.
This section first describes the annotation procedure (§\ref{sec:pilot-annot-proc}) and then reviews the quality and effectiveness of the annotation process (§\ref{sec:pilot-analysis}).

\subsection{Annotation Procedure}
\label{sec:pilot-annot-proc}
For the user study, we hired three non-expert annotators, all native English speakers and either first-degree students or graduates. They underwent three training iterations, each on a single document. 
Annotators were instructed to follow the methodology used in creating the MATRES dataset \cite{ning-etal-2018-multi} for selecting events and annotating temporal relations. This approach was chosen for its simplicity, making it suitable for non-expert annotators.
MATRES distinguishes event mentions by their temporal characteristics --- events that are actual or ``anchorable in time'' (e.g., they \textit{\underline{won}} the game) are included, while wishful, intentional, or conditional events (e.g., I wish they \textit{\underline{win}} the game) are excluded. Temporal relations are determined only based on the starting times of the events. For coreference and causal relations, annotators adhere to the tool's workflow (§\ref{sec:tool-flow}).


The annotation process was conducted on six news documents, each approximately 500 words long, where each document was annotated by all three annotators for measuring agreement. To extract the initial set of event mentions, we applied the event detection method proposed by \citet{cattan-etal-2021-cross-document}, which identified an average of 60 event mentions per document. Annotators then identified ``anchorable'' event mentions using the tool's event selection step, averaging 35 events per document. To further refine the set and manage the study’s scope, we followed prior approaches (§\ref{sec:background-relations}) by instructing annotators to select the 16–18 most salient events.
Following this selection, annotators proceeded to annotate the three event relation types.
Over this process, we measure inter-annotator agreement, annotation time, and the number of annotation steps for each task.

\input{tables/agree_tab}

\subsection{Analysis}
\label{sec:pilot-analysis}
The final annotated set included 102 event mentions over the 6 documents, averaging 17 mentions per document, resulting in 816 distinct annotated pairs of event mentions, with an average of 136 mention pairs per document. In comparison, the prominent MATRES temporal relation dataset \cite{ning-etal-2018-multi} contains an average of 22 event mentions per document but only 50 annotated mention pairs per document.

Notably, as shown in Table~\ref{tab:annot_agree}, agreement between annotators is high across all three relations and comparable to the recent MAVEN-ERE dataset, indicating that the pilot reliably reflects a data annotation process on par with other datasets.

We also observe in Table~\ref{tab:annot_time} that the annotation of temporal relations roughly takes 44 minutes to complete, which is significantly more demanding than for the other two types of relations. 
This finding is reasonable, as temporal relation annotation requires classifying each relation into one of four classes.
In contrast, coreference and causal relations involve identifying connections within a set of events relative to a focal event and its temporal context (\textit{equal} for coreference and \textit{after} for causal). 
Importantly, Table~\ref{tab:annot_step} indicates that \App{} assists in significantly reducing complexity for all three relation types.


%% file: tables/agree_tab.tex
\begin{table}[!t]
    \centering
    \resizebox{0.48\textwidth}{!}{
    \begin{tabular}{@{}l|ccc@{}}
        \toprule
        & \makecell{Temporal($\kappa$)} & \makecell{Coreference($B^3$)} & \makecell{Causal($\kappa$)} \\
        \midrule
        A and B & 0.72 & 0.98 & 0.83 \\
        A and C & 0.75 & 0.93 & 0.77 \\
        B and C & 0.68 & 0.95 & 0.74 \\
        \midrule
        Average (\App{}) & 0.72 & 0.96 & 0.78 \\
        \midrule
        \midrule
        MAVEN-ERE & 0.68 & 0.91 & 0.7 \\
        \bottomrule
    \end{tabular}}
    \caption{\textbf{Agreement between Annotators:} For temporal and causal relations, the \textit{kappa} coefficient \cite{kappa-1973} was used, calculated using scikit-learn \cite{buitinck2013apidesignmachinelearning}. For coreference, the \textit{B-Cubed} F1 score \cite{bagga-baldwin-1998-entity-based} was applied. A/B/C represent the three annotators. Additionally, for comparability, we report the agreement values from the recent MAVEN-ERE dataset \cite{wang-etal-2022-maven}.}
    \label{tab:annot_agree}
\end{table}

%% file: tables/annot_time.tex
\begin{table}[!t]
    \centering
    \resizebox{0.48\textwidth}{!}{
    \begin{tabular}{@{}l|ccc|c@{}}
        \toprule
        & \makecell{A} & \makecell{B} & \makecell{C} & \makecell{Average Time} \\
        \midrule
        Temporal & 45 & 38 & 50 & 44.3 \\
        Coreference & 8 & 5 & 10 & 7.7 \\
        Causal & 16 & 15 & 21 & 17.3 \\
        \midrule
        Total Time (min) & 69 & 58 & 81 & 69.3 \\
        \bottomrule
    \end{tabular}}
    \caption{\textbf{Annotation Time (in minutes):} The average time taken by each annotator to complete each task for a single document. A/B/C represent the three annotators.}
    \label{tab:annot_time}
\end{table}


%% file: tables/annot_steps.tex


\begin{table}[!t]
\resizebox{\columnwidth}{!}{%
\begin{tabular}{l|ccc|c}
\hline
Relation Types & A    & B    & C    & \makecell{Average \\ Reduction} \\ \hline
Temporal       & 56.7 & 54.6 & 65.6 & 56\%              \\
Coreference    & 4.5  & 4.5  & 4.8  & 96\%              \\
Causal         & 79.2 & 79.2 & 79.2 & 41\%              \\ \hline
\end{tabular}%
}
\caption{\textbf{Annotation Steps Made:} For each document, we calculated the average number of pairs requiring classification to obtain a complete annotation for all event mention pairs (excluding symmetric ones), which averaged 136 pairs per document. We then measured the average number of pairs per document that \App{} presented for judgement to each annotator (A/B/C). The \textbf{Average Reduction} represents the percentage by which \App{} reduced the number of pairs from the total average of 136 pairs per document.}
\label{tab:annot_step}
\end{table}

%% file: sections/conclusion.tex
\section{Conclusion}

In this paper we introduced \App{}, a novel tool for the end-to-end annotation of \textit{complete} and \textit{consistent} event relation over targeted events in a text, integrating temporal, coreference, and causal relations into a unified workflow. Starting with temporal relations, we leverage their properties to reduce the manual annotation workload and ensure consistency. Annotating temporal relations first allows the coreference step to focus only on \textit{equal} relations and the causal step on \textit{before} relations, thereby reducing the set of pairs requiring consideration in each step. Additionally, annotating coreference before causal relations enables all coreferring event mentions to be treated as a single event, further reducing annotation complexity. Last, we assessed the utility of \App{} through a pilot study, measuring annotation complexity, data quality via inter-annotator agreement (IAA), and annotation time efficiency. Results demonstrate that \App{} significantly reduces annotation complexity while maintaining high IAA. We hope that \App{} will facilitate the creation of new datasets that complement existing resources, focusing on constructing complete event relation annotation for targeted events in a text.

%% file: sections/appendix.tex

\section{Causal Scheme Experiment}
\label{app:cause-scheme-expr}
To address the complexity of causal annotation, we experimented with various annotation schemes, ultimately focusing on RED \cite{ogorman-etal-2016-richer} and EventStoryLine \cite{caselli-vossen-2017-event} methodologies. Our findings showed that annotating causality using the RED methodology required more than twice the time compared to the EventStoryLine methodology. This difference is attributed to two main factors: (1) RED requires evaluating all pairs of events, whereas EventStoryLine involves a single pass, evaluating each event with all its preceding events; (2) RED demands assessing whether the target event was ``inevitable'' given the source event for each pair. We observed that annotators often struggled with the concept of inevitability, even when clear causal indicators were present in the text. For instance, in the sentence, ``He answered the phone \emph{because} it rang,'' annotators found it challenging to evaluate causality. This led to low inter-annotator agreement, around 0.2 \textit{kappa} (slight to fair agreement). In contrast, using the EventStoryLine methodology, annotators achieved a 0.78 \textit{kappa}, indicating substantial agreement.

\section{Transitive Closure}
\label{appx:transitive-algo}
We devised a straightforward algorithm designed to infer transitive relations. Our algorithm operates in three steps: (1) We begin with a matrix \(X \in M_{n \times n}\) (\(n\) being the number of events), where each cell in the matrix represent a pair and indicates whether a direct relation has been annotated between them. (2) For every annotation made, we utilize Warsheall's algorithm \cite{warsheall-1962} to compute transitive closures, adapting it to accommodate the unique transitive characteristics of each relation type (as detailed in Table~\ref{tab:disc_temp}). The output of the algorithm is a matrix \(\hat{X} \in M_{n \times n}\), where both direct relations and inferred transitive relations are marked. (3) The tool then examines \(\hat{X}\) (excluding its diagonal) to identify the remaining pairs requiring annotation by collecting all pairs whose corresponding cells are assigned the \(ANNOTATE\) relation (Table~\ref{tab:disc_temp}).


\section{Prioritization Strategy}
\label{appx:prior-strategy}


Our prioritization strategy is designed to increase the likelihood of utilizing the transitive closure algorithm for automatically detecting transitive relations \ref{appx:transitive-algo}. It is inspired by two key observations: (1) the \textit{before} relation is the most frequent in prior temporal datasets \cite{kishore-he-2024-unveiling}, and (2) story narratives naturally tend to progress chronologically as they unfold.

To that end, after applying the transitive closure algorithm (Appendix~\ref{appx:transitive-algo}), all pairs requiring annotation (i.e., marked with \(ANNOTATE\)) are arranged in the order of their appearance in the text. The first pair requiring annotation is selected, followed by examining the next pair requiring annotation. Finally, the last pair that shares the second node with this pair is selected as the next candidate pair to present to the annotator. This simple approach simulates a depth-first-like search, prioritizing yet unreached mentions. For example, in Figure~\ref{fig:example}, once the pair \textit{`accident-collided'} is annotated as \textit{EQUAL}, the next unhandled pair is \textit{`accident-damage'}, and the last pair sharing the second node `damage' is \textit{`collided-damage'}.

\section{Consistency Checking Algorithm}
\label{appx:discrepancy}
To manage potential discrepancies, we incorporated rule-based checks into the transitive closure algorithm (Appendix~\ref{appx:transitive-algo}), similar to the transitive constraint logic proposed by \cite{ning-etal-2018-joint}. These checks ensure that no direct annotation contradicts a newly identified transitive relation (or vice versa).
At the base of our discrepancy algorithm, three nodes are considered: \(i\), \(j\), and \(k\), where \(i\) and \(j\) represent our target nodes, and \(k\) represents any node through which there is a path from \(i\) to \(j\). This logic then examine whether the direct relation \(\{i, j\}\) contradicts the temporal relation inferred from combining the two edges that form the path (the rules are shown in Table~\ref{tab:disc_temp}). If the direct relation \(\{i, j\}\) contradicts the inferred relation \(\{i, k, j\}\), \App{} will alert the annotator to resolve the contradiction. For the coreference relation, we handle cases where a user assigns a mention to one event cluster and later assigns the same mention to a different cluster. In such instances, the tool notifies the annotator about the misalignment and prompts them to verify their selection. If the annotator confirms the change, the event mention is removed from its current cluster and added to the new one.

\section{Temporal Constraints}
\label{appx:temporal-constraints}
Table~\ref{tab:disc_temp} presents the temporal constraints employed in our transitivity and discrepancy detection algorithms.

\input{tables/disc_temp}

\section{Printscreens of \App{}}
\label{appx:additional_printscreens}
\begin{figure*}[t!]
\centering
\includegraphics[width=\textwidth]{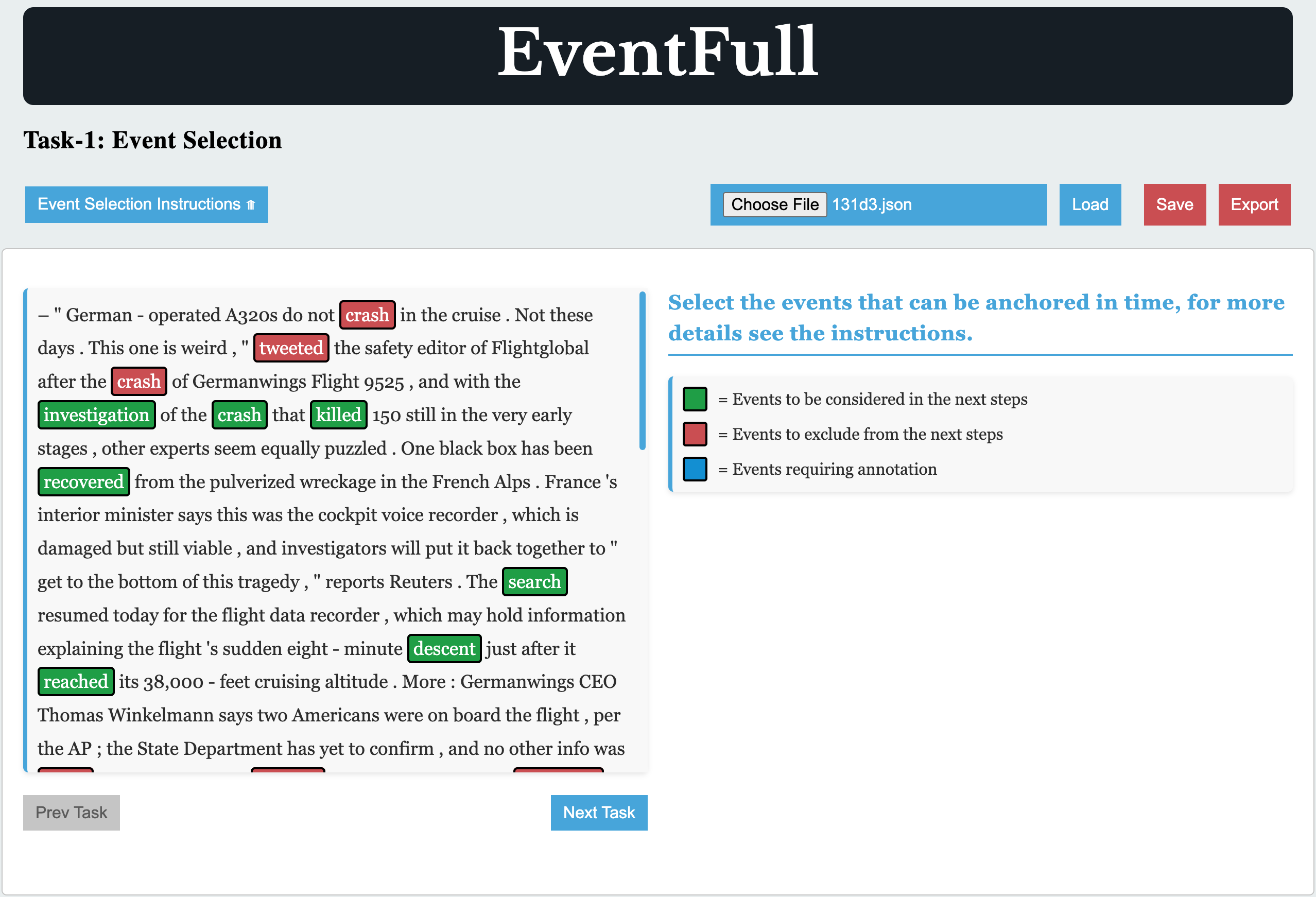}
\caption{\textbf{Event Selection Annotation Step:} This optional step aims to refine the set of events (detailed in §\ref{sec:tool-flow}) by selecting the events to be considered in subsequent steps. Annotators can access guidelines by clicking the ``Event Selection Instruction'' button. After categorizing all events as either \colorbox{teal}{\textcolor{white}{event}} or \colorbox{purple}{\textcolor{white}{no-event}}, they can proceed to the next annotation task by clicking the ``Next Task'' button.}
\label{fig-appx:select}
\end{figure*}

\begin{figure*}[t!]
\centering
\includegraphics[width=\textwidth]{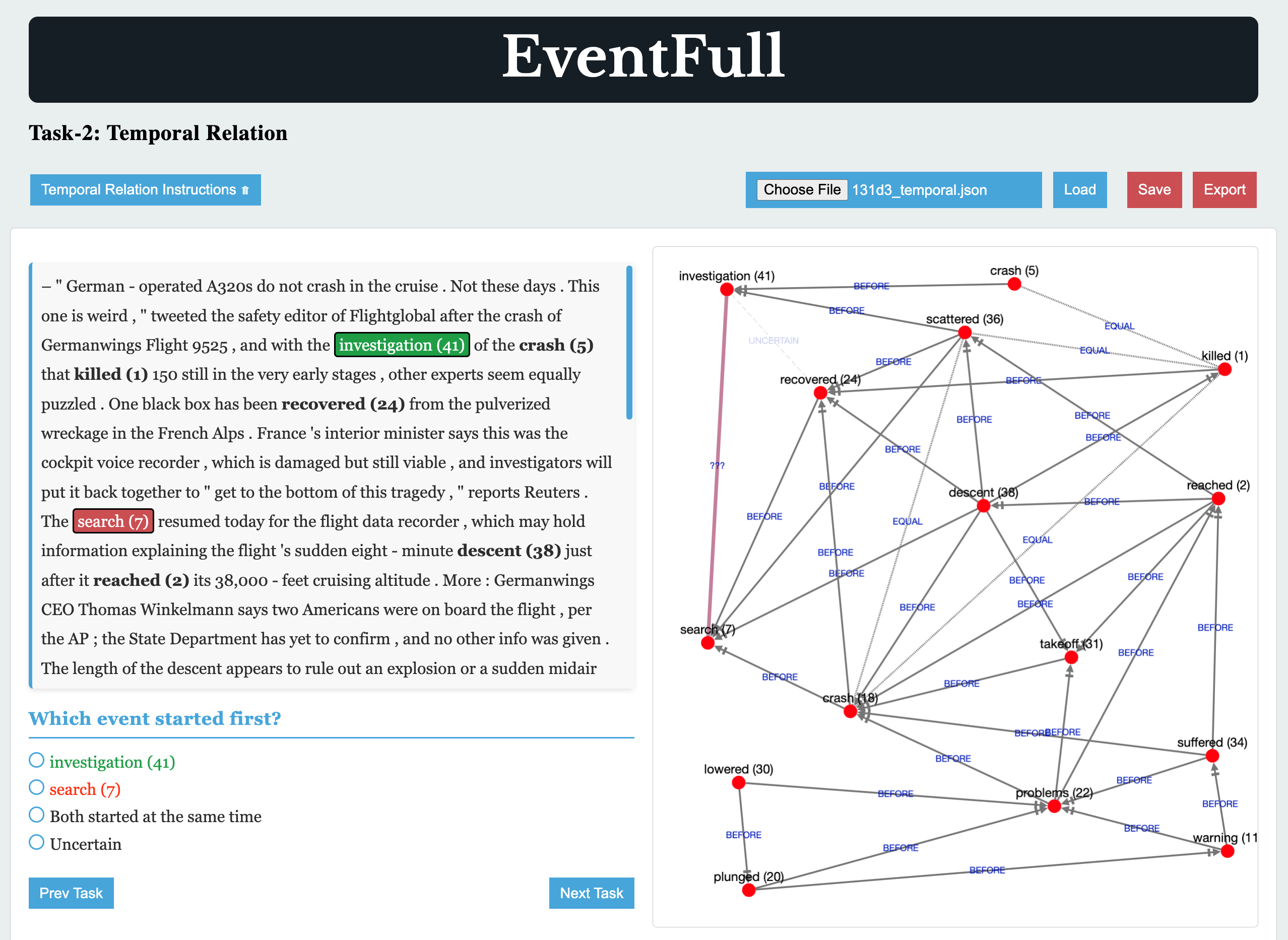}
\caption{\textbf{Temporal Relation Annotation Step:} Annotators determine the temporal relations for each candidate pair based on the starting point of the events by selecting the appropriate radio button option (detailed in §\ref{sec:tool-flow}). Events requiring annotation are highlighted in \colorbox{teal}{\textcolor{white}{green}} and \colorbox{purple}{\textcolor{white}{red}} within the context, with the selected relation displayed in \textcolor{red}{red} in the visualization graph. All other event mentions are marked in bold within the context. The graph visualization tracks annotators' progress and allows manual selection by clicking on two nodes representing events in the graph. Annotated relations are displayed in the graph with their corresponding temporal label. Each event mention is assigned an identifier in parentheses, visible in both the context and graph, to facilitate locating events.}
\label{fig-appx:temporal}
\end{figure*}

\begin{figure*}[t!]
\centering
\includegraphics[width=\textwidth]{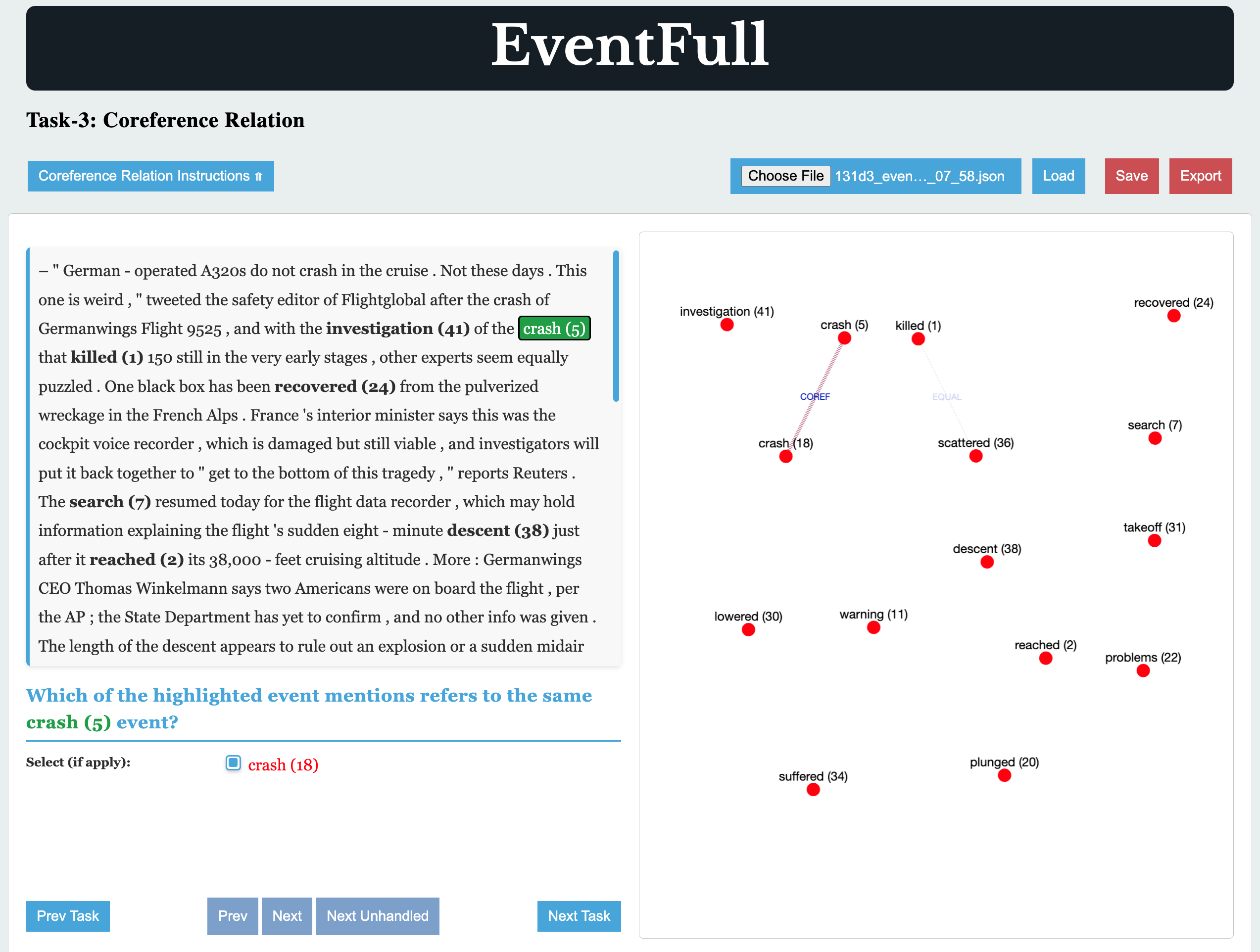}
\caption{\textbf{Coreference Relation Annotation Step:} Annotators determine coreference relations among all candidates annotated in the \textit{temporal} step as having an \textit{equal} time relation. The event mention representing the event cluster is highlighted in \colorbox{teal}{\textcolor{white}{green}}, while all candidate event mentions (sharing an equal time with it) are highlighted in \colorbox{purple}{\textcolor{white}{red}}. Annotators are required to check the checkbox of the \textcolor{red}{red} event mentions that corefer with the green one and proceed to the next group using the ``Next Unhandled'' button. A graph visualization tracks annotators' progress, displaying only relevant relations for this step (i.e., equal and coreference). The relations under scrutiny are highlighted in \textcolor{red}{red}, while not-corefer relations appear faded.}
\label{fig-appx:coref}
\end{figure*}

\begin{figure*}[t!]
\centering
\includegraphics[width=\textwidth]{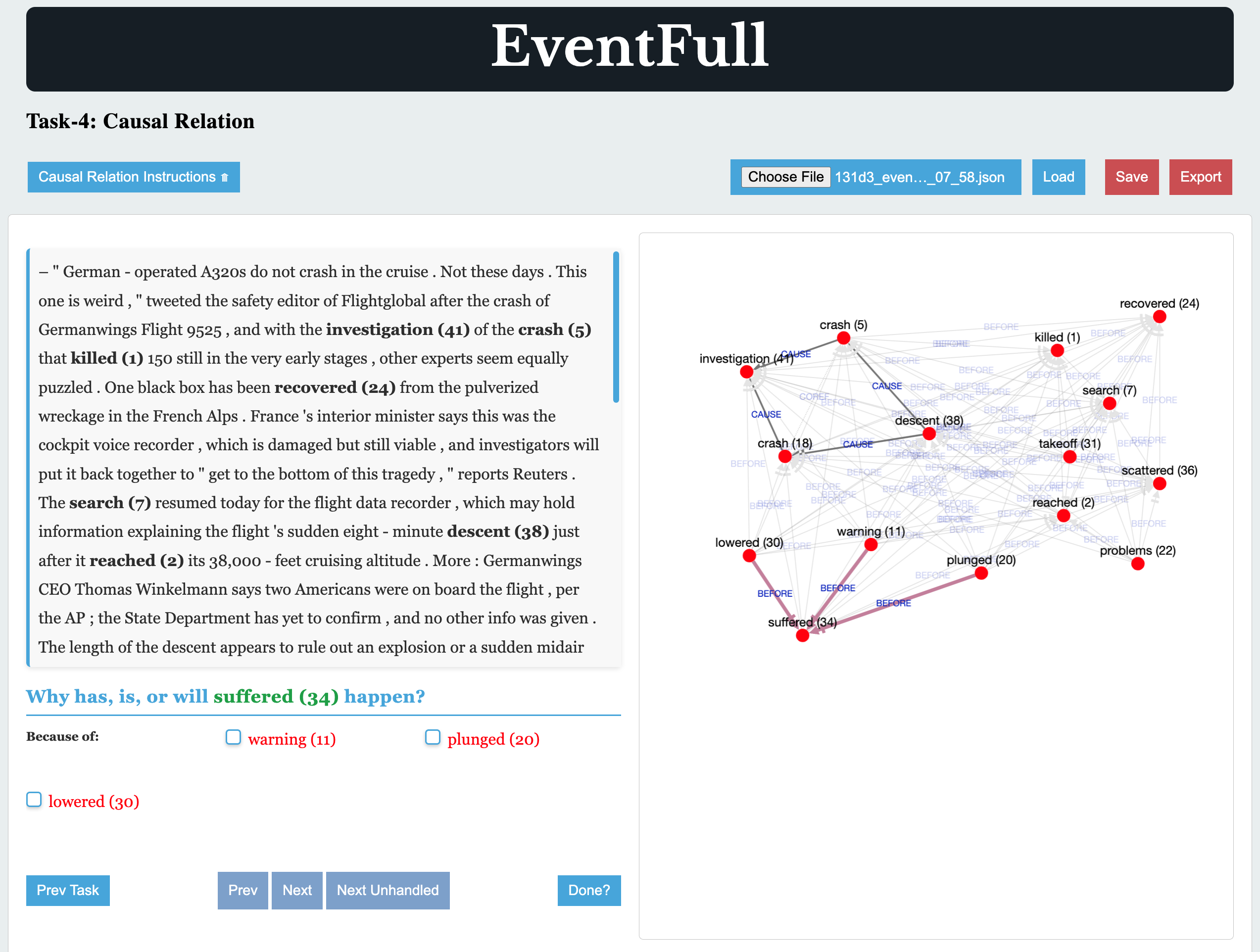}
\caption{\textbf{Causal Relation Annotation Step:} Annotators determine causal relations among all candidates annotated in the \textit{temporal} step as having a \textit{before} time relation. The event mention in focus is highlighted in \colorbox{teal}{\textcolor{white}{green}}, while all preceding event mentions are highlighted in \colorbox{purple}{\textcolor{white}{red}}. Annotators are required to check the checkbox of the \textcolor{red}{red} event mention(s) that caused the green one to occur and proceed to the next group using the ``Next Unhandled'' button. A graph visualization tracks annotators' progress, displaying only relevant relations for this step (i.e., BEFORE and CAUSE). The relations under scrutiny are highlighted in \textcolor{red}{red}, with all relations already annotated as CAUSE clearly visible, while non-causal relations appear faded.}
\label{fig-appx:causal}
\end{figure*}

We present examples of \App{} (Figures~\ref{fig-appx:select}, \ref{fig-appx:temporal}, \ref{fig-appx:coref}, and \ref{fig-appx:causal}), showcasing the four annotation steps.

\section{\App{} Annotation Guidelines}
\label{appx:guidelines}
\begin{figure*}[t!]
\centering
\includegraphics[width=\textwidth]{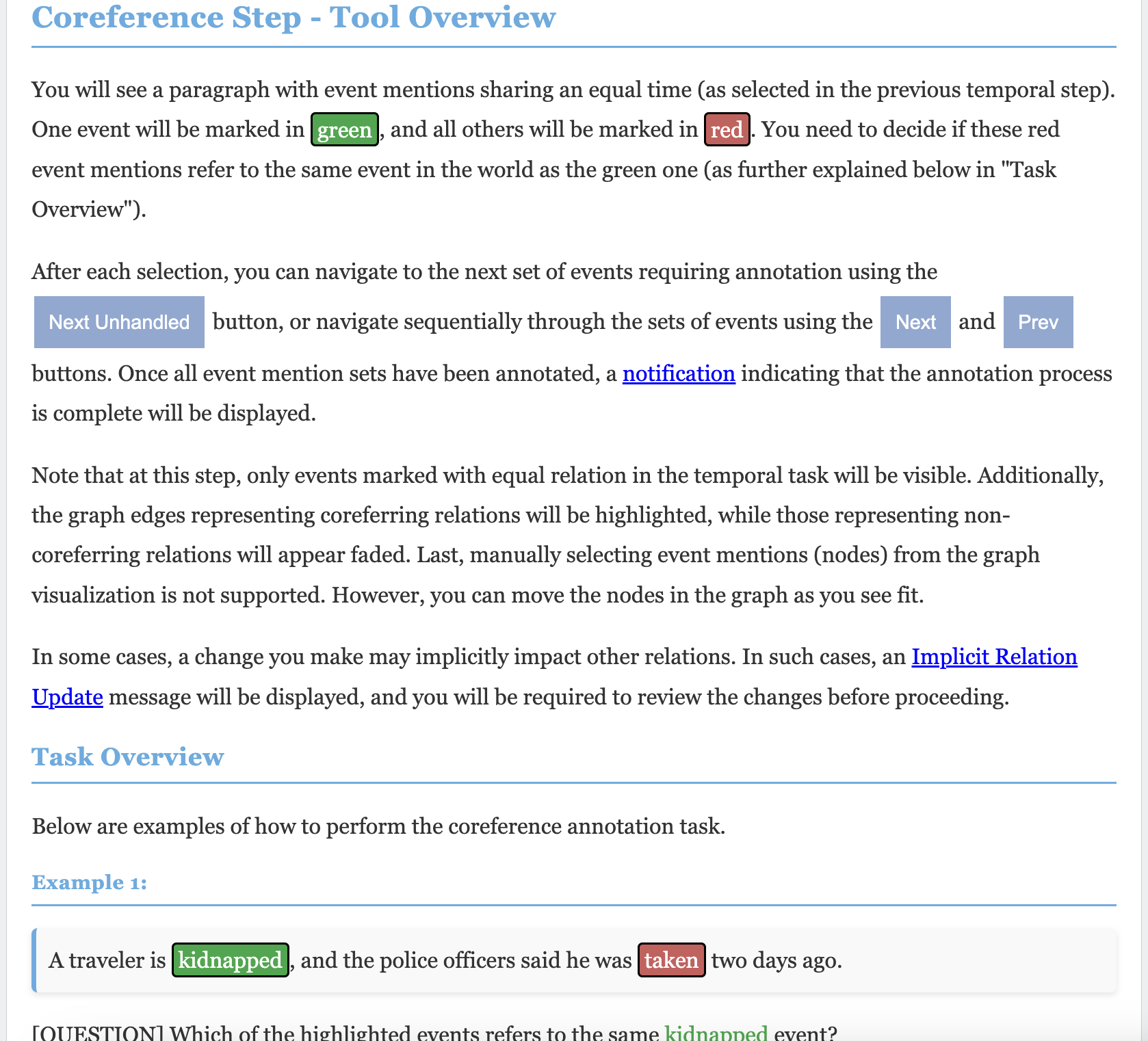}
\caption{An example of the guidelines used in the pilot study, accessible by clicking the ``Coreference Relation Instructions'' button in the Coreference annotation step. Similarly, guidelines are available for the other steps within the tool, allowing annotators to revisit them as needed while performing the annotation task.}
\label{fig-appx:guidelines}
\end{figure*}

The guidelines in \App{} are fully customizable, allowing anyone managing an annotation task to edit them to align with any annotation scheme. We present an example of the guidelines we used in our pilot study, accessible through the tool's UI (Figures~\ref{fig-appx:guidelines}).

%% file: tables/disc_temp.tex
\begin{table}[!t]
    \centering
    \resizebox{0.35\textwidth}{!}{
    \begin{tabular}{@{}ccccc@{}}
        \toprule
        \makecell{} &\makecell{\(\{i, k\}\)} & \makecell{\(\{k, j\}\)} & \makecell{\(\{i, j\}\)} \\
        \midrule
        1 & BEFORE & BEFORE & BEFORE \\
        2 & BEFORE & EQUAL & BEFORE \\
        3 & EQUAL & BEFORE & BEFORE \\
        \midrule
        4 & AFTER & AFTER & AFTER \\
        5 & AFTER & EQUAL & AFTER \\
        6 & EQUAL & AFTER & AFTER \\
        \midrule
        7 & EQUAL & EQUAL & EQUAL \\
        \midrule
        8 & BEFORE & AFTER & ANNOTATE \\
        9 & AFTER & BEFORE & ANNOTATE \\
        10 & VAGUE & VAGUE & ANNOTATE \\
        11 & EQUAL & VAGUE & ANNOTATE \\
        12 & VAGUE & EQUAL & ANNOTATE \\
        13 & BEFORE & VAGUE & ANNOTATE \\
        14 & VAGUE & BEFORE & ANNOTATE \\
        15 & AFTER & VAGUE & ANNOTATE \\
        16 & VAGUE & AFTER & ANNOTATE \\
        \bottomrule
    \end{tabular}}
    \caption{\(\{i, k\}\) and \(\{k, j\}\) represent annotated paths between events \(i\) to \(k\), and \(k\) to \(j\). \(\{i, j\}\) represents the inferred transitive relation between events \(i\) and \(j\) via \(k\). The ANNOTATE relation indicates that this relation should be presented to the annotator for verification.}
    \label{tab:disc_temp}
\end{table}